\documentclass[11pt]{article}
\usepackage{coling2020}
\usepackage{times}
\usepackage{url}
\usepackage{latexsym}
\usepackage{multirow}
\usepackage{subcaption}
\usepackage{graphicx}
\usepackage{color}

\usepackage{amsmath}
\usepackage{amssymb}
\usepackage{amsthm}

\usepackage{tikz}
\usetikzlibrary{graphs,arrows,positioning,trees,decorations.pathmorphing,chains,shapes,decorations.pathreplacing,quotes,calc}
\tikzstyle{hidden} = [circle,draw]
\tikzstyle{visible} = [circle,draw,fill=black!10]
\tikzstyle{internal} = [circle,draw,fill=black!10]
\tikzstyle{selected} = [line width=3pt, draw=myRed]
\definecolor{myBlue}{RGB}{0,117,189}
\definecolor{myRed}{RGB}{217,83,25}
\definecolor{myYellow}{RGB}{237,177,32}
\definecolor{myGreen}{RGB}{83,217,25}
\colingfinalcopy 

\newcommand{\Tensor}[1]{\ensuremath{\mathbf{ #1}}}

\DeclareMathOperator*{\argmax}{arg\,max}

\newcommand{\rulesep}{\unskip\ \vrule\ }

\title{Learning from Non-Binary Constituency Trees via Tensor Decomposition}

\author{Daniele Castellana \\
  University of Pisa, Pisa \\
  {\tt daniele.castellana@di.unipi.it} \\\And
  Davide Bacciu \\
  University of Pisa, Pisa \\
  {\tt bacciu@di.unipi.it} \\}

\date{}

\begin{document}
\maketitle
\begin{abstract}
Processing sentence constituency trees in binarised form is a common and popular approach in literature. However, constituency trees are non-binary by nature. The binarisation procedure changes deeply the structure, furthering constituents that instead are close. In this work, we introduce a new approach to deal with non-binary constituency trees which leverages tensor-based models. In particular, we show how a powerful composition function based on the canonical tensor decomposition can exploit such a rich structure. A key point of our approach is the weight sharing constraint imposed on the factor matrices, which allows limiting the number of model parameters. Finally, we introduce a Tree-LSTM model which takes advantage of this composition function and we experimentally assess its performance on different NLP tasks.
\end{abstract}

\section{Introduction}

\blfootnote{This work is licensed under a Creative Commons Attribution 4.0 International Licence. Licence details: http://creativecommons.org/licenses/by/4.0/.}

One of the fundamental problems in Natural Language Processing (NLP) is learning a distributed encoding of sentences, as this is the stepping stone for many NLP tasks, such as sentence classification, sentiment analysis and natural language inference. The multitude of approaches addressing this problem can be categorised according to how a sentence is represented.

The simpler sentence representation is bag-of-words, which depicts sentences as words multisets ignoring the word order. Despite the simple representation, it has been used to obtain meaningful sentence encodings \cite{Iyyer2015,Wieting2016,Arora2017}. 

Sequence representation overcomes this limitation considering the sentence as an ordered sequence of words. It allows building models which progressively constructs a sentence encoding, processing one word at the time. Recurrent Neural Network \cite{Elman1990} and Long-Short Term Memory (LSTM) \cite{Hochreiter1997} are probably the most famous models which use this representation.

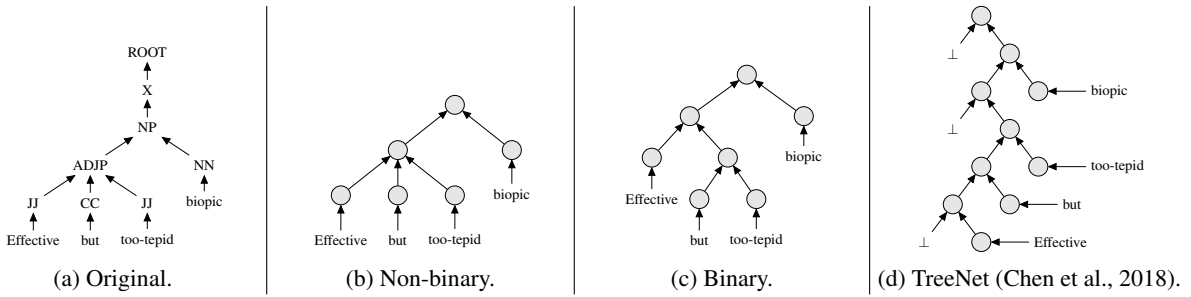
\begin{figure}
\begin{subfigure}[b]{0.24\textwidth}
        \centering
        \scalebox{0.5}{
        \begin{tikzpicture}[edge from parent/.style={draw,<-,>=triangle 45}, 
            every node/.style = {minimum size = 0.5cm},
            level 1/.style={level distance=1cm},
            level 3/.style={sibling distance=3cm},level 4/.style={sibling distance=1.5cm}]
            \node [minimum size=0cm] (qu) {ROOT}
            child { 
            	node {X} 
            	child {
            		node {NP}
            		child{
            			node[] {ADJP}
            			child{
            				node {JJ}
            				child{
            					node {Effective}
            				}
            			}
            			child{
            				node {CC}
            				child{
            					node {but}
            				}
            			}
            			child{
            				node {JJ}
            				child{
            					node {too-tepid}
            				}
            			}
            		}
            		child{
            			node {NN}
            			child {
            				node {biopic}
            			}
            		}
            	}
            };
            \end{tikzpicture}}
\caption{Original.}\label{fig:orig_tree}\end{subfigure}
\rulesep
\begin{subfigure}[b]{0.24\textwidth}
        \centering
        \scalebox{0.5}{
        \begin{tikzpicture}[edge from parent/.style={draw,<-,>=triangle 45}, 
            every node/.style = {minimum size = 0.5cm},
            level 1/.style={sibling distance=3cm, level distance=1.2cm},
            level 2/.style={sibling distance=1.5cm}]
            \node [visible] (qu) {}
            child { 
            	node[visible] {}
            	child{
            	    node[visible] {}
            	    child{
            		    node {Effective}
        		    }
            	}
            	child{
            	    node[visible] {}
            		child{
            		    node {but}
        		    }
            	}
            	child{
            	    node[visible] {}
            		child{
            		    node {too-tepid}
        		    }
            	}
            }
            child{
        	    node[visible] {}
        		child{
        		    node {biopic}
    		    }
            };
            \end{tikzpicture}}
\caption{Non-binary.}\label{fig:const_tree}\end{subfigure}
\rulesep
\begin{subfigure}[b]{0.23\textwidth}
        \centering
        \scalebox{0.5}{
        \begin{tikzpicture}[edge from parent/.style={draw,<-,>=triangle 45}, 
            every node/.style = {minimum size = 0.5cm},
            level 1/.style={sibling distance=3cm, level distance=1.1cm},
            level 2/.style={sibling distance=2cm},level 3/.style={sibling distance=1.5cm}]
            \node [visible] (qu) {}
            child { 
            	node[visible] {}
            	child{
	            	node[visible] {}
            	    child{
            		    node {Effective}
        		    }
            	}
            	child{
                	node[visible] {}
                	child{
                	    node[visible] {}
                		child{
                		    node {but}
            		    }
        		    }
                	child{
                	    node[visible] {}
                		child{
                		    node {too-tepid}
            		    }
                	}
            	}
            }
            child{
            	node[visible] {}
        		child{
        		    node {biopic}
    		    }
            };
            \end{tikzpicture}}
\caption{Binary.}\label{fig:bin_const_tree}\end{subfigure}
\rulesep
\begin{subfigure}[b]{0.25\textwidth}
        \centering
        \scalebox{0.5}{
        \begin{tikzpicture}[edge from parent/.style={draw,<-,>=triangle 45}, 
            every node/.style = {minimum size = 0.5cm},
            level 1/.style={level distance=1cm},
            level 3/.style={sibling distance=1.5cm},level 4/.style={sibling distance=1.5cm}]
		\node [visible] (qu) {}
		child{
			node {$\bot$}            			
		}
        child { 
        		node[visible] {}
        		child{
        			node[visible] {}  	
        			child{
        					node {$\bot$}            			
        				}            			
        			child{
        				node[visible] {}  			
        				child{
        					node[visible] {}  	
					        child{
            					node[visible] {}  		
	            				child{
	            					node {$\bot$}            			
	            				}    	 	            					
	            				child{
	            					node[visible](ef) {}
	            					child{
	            					    node[right=of ef] {Effective}            			
            					    }
	            				}          
            				}
            				child{
            					node[visible](but) {}
            					child{
            					    node[right=of but] {but}            			
            					}
            				}            					
        				}
        				child{
        					node[visible](too) {}
        					child{
        					    node[right=of too] {too-tepid}            			
    					    }
        				}
        			}
        		}
        		child{
        			node[visible](bio) {}
        			child{
        			    node[right=of bio] {biopic}            			
        			}
        		}
    	};
\end{tikzpicture}}
\caption{TreeNet \cite{Chen2018}.}\label{fig:tree_net_tree}\end{subfigure}
\caption{Constituency tree of sentence "\emph{Effective but too-tepid biopic}" taken from SST \cite{Socher2013} test set.}
\label{fig:const_tree_example}
\end{figure}

A key aspect of sentences, which is missing in sequential processing, is compositionality. For example, the sentence "\emph{The sky is blue and the grass is green}" is obtained by composing the two sub-phrases  "\emph{The sky is blue}" and "\emph{the grass is green}" with the conjunction "\emph{and}". The intrinsic compositionality of sentences makes them suitable for a tree representation, where the whole sentence (the root) is built in terms of sub-phrases (the internal nodes) which in turn are defined in terms of smaller constituents; the base cases are words (the leaves) since they are the atomic piece of information. This representation takes the name of \emph{constituency tree}. In Fig.\ \ref{fig:orig_tree} we show the constituency tree of the sentence "\emph{Effective but too-tepid biopic}": the leaves are the words while internal nodes represent syntactic categories which are the constituents of the whole sentence.

There are many models which compute a sentence encoding starting from its constituency tree. For our purposes, we restrict the discussion on bottom-up Recursive Neural Networks (RecNNs) \cite{Goller1996,Frasconi1998}. The parsing direction is constrained by the structure of constituency trees, having information (i.e. words) on leaf nodes. In this domain, we refer to the term \emph{composition function} to indicate the state-transition function which computes the \emph{representation} (i.e. the hidden state) of a tree node combining the representation of its \emph{constituents} (i.e. the hidden state of its child nodes). Then, the hidden state of the root (i.e. the whole sentence) is taken as sentence encoding.

The Matrix-Vector Recurrent Neural Network (MV-RNN) \cite{Socher2012} and the Recursive Neural Tensor Network (RNTN) \cite{Socher2013} apply the RecNN architecture to binary constituency trees using complex composition functions. 

\newcite{Tai2015} extends the well known Long-Short Term Memory \cite{Hochreiter1997} architecture to tree-structured data. They propose two different Tree-LSTMs (that we discuss in Section \ref{sec:treeLSTM}): the $N$-ary Tree-LSTM defines a composition function which considers constituent order while the child-sum Tree-LSTM ignores such an order. However, only the former model is applied to binary constituency trees. The latter is applied to dependency trees, which are another kind of tree representation for sentences, out of our scope.

In recent years, Tree-LSTM has been used as a building block to develop more sophisticated models. For example, \newcite{Huang2017}, \newcite{Liu2017b}, \newcite{Kim2019},  \newcite{Shen2020} build new Tree-LSTM models which define dynamic composition functions depending on syntactic categories (i.e. Part-Of-Speech tags). Instead, \newcite{Teng2017} introduces a Bidirectional Tree-LSTM which takes advantage of both parsing directions: bottom-up and top-down. As we stated before, constituency trees are intrinsically bottom-up; to this end, the author introduces a first bottom-up pass, called \emph{head lexicalization}, to propagate information from leaves to the root. All these models are applied only to binary constituency trees.

Thus far, we have shown that most of the models compute sentences encodings starting from binary constituency trees. This simplification solves one crucial problem of tree-structured data: the variable number of child nodes. However, the price to pay is the loss of structural information. For example, in Fig.\ \ref{fig:const_tree} and Fig.\ \ref{fig:bin_const_tree} we report the constituency and the binary constituency tree of the sentence "\emph{Effective but too-tepid biopic}". Comparing the two representation, we can observe that binary tree has one more node that breaks the ternary relation in the non-binary tree; in general, to break a node with $L$ child nodes, we need to add $L-2$ new nodes. All these new nodes create a chain which moves away the child nodes of the n-ary relation from their parent. The composition of them is obtained by considering one child at a time, as it happens in sequence representation. Hence, the binarisation removes the equality among child nodes, with the risk of weakening contribution of child nodes that are moved far away from their parent and strengthening the contribution of the ones that remain close.

As far as we know, the only work which builds a model suitable for non-binary constituency trees is the TreeNet \cite{Cheng2018}. The idea is to consider all child nodes in a chain: the hidden state of a node depends on the hidden state of its left sibling and its rightmost child. Even if the model itself works with non-binary trees, the composition function expressed is binary since it always composes two elements. We discuss this observation in details in Sec. \ref{eq:treeNet}.

The definition of models for non-binary constituency trees requires to go beyond the standard definition of composition function. Standard RecNNs define learnable composition functions which are based on the summation of the contribution of each constituent. \newcite{Castellana2020} proposed a generalisation of such sum-based composition functions leveraging more expressive multi-affine maps represented as tensors. The exponential number of parameters with respect to the tree out-degree (i.e. the maximum number of children for each node in the tree) required by the full-tensorial approach can be controlled by applying tensor decomposition. The tensorial models outperform sum-based models, especially when the tree out-degree increases \cite{Castellana2020,Castellana2020esann}.

Within the scope of this paper, we unveil that non-binary constituency trees can be effectively exploited to improve predictive performance in NLP task, showing that more powerful composition functions are necessary to take advantages of such a rich representation. To this end, we introduce two new Tree-LSTM models which leverage canonical tensor decomposition: the former is suitable for binarised constituency trees, while the latter can process general non-binary constituency trees imposing weight sharing on the tensor decomposition factors. Finally, we test the quality of sentence encodings produced by our models on different NLP tasks, showing that the combination of a rich representation and a powerful composition function is able to outperform baseline models using the same number of parameters.

\section{Related Models}
In this section, we discuss three architectures from the literature that are related to the approach put forward in this paper and are used as baselines in our empirical analysis.

\subsection{Tree-Structured LSTM} \label{sec:treeLSTM}
The Long-Short Term Memory (LSTM) \cite{Hochreiter1997} is one of the most popular neural architecture to process sequential data. \newcite{Tai2015} proposed two extensions of such architecture to handle tree-structured information: the $L$-ary Tree-LSTM and the Child-Sum Tree-LSTM. Both models propagate the information along the input tree structure in a bottom-up fashion (i.e. from the leaves to the root). Hence, at each node, the tree-LSTM cell aggregates the hidden child states to compute its gates leveraging sum-based composition functions. 

\paragraph{$L$-ary Tree-LSTM} \cite{Tai2015} assumes that child nodes are ordered and then they can be indexed from $1$ to $L$, where $L$ denotes the maximum node output degree, i.e.\ the maximum number of children a node can have. Let $v$ a generic node, we indicate with $h_{vj} \in \mathbb{R}^d$ and $c_{vj} \in \mathbb{R}^d$ respectively the hidden state and the memory cell of its $j$-th child node; hence, the $L$-ary Tree-LSTM cell computation is defined by the following equations:
\begin{equation}
    \begin{alignedat}{2}
        t_v &= \sigma\left( W^tx_v + \sum_{j=1}^{L}U^t_j h_{vj} + b^t\right) \quad t \in \{i,o,u\}, \quad & \quad c_v &= i_v \odot u_v + \sum_{j=1}^{L} f_{vj} \odot c_{vj},\\
        f_{vk} &= \sigma\left( W^fx_v + \sum_{j=1}^{L}U^f_{kj} h_{vj} + b^f_{k}\right) k \in [1,L],  \quad & \quad h_v &= o_v \odot \text{tanh}(c_v),
    \end{alignedat}
    \label{eq:laryLSTM}
\end{equation}
where $c_v \in \mathbb{R}^d$, $h_v \in \mathbb{R}^d$ and $x_v \in \mathbb{R}^n$ are the memory cell state, the hidden state and the label of node $v$; $\{i_v, o_v, u_v\} \in \mathbb{R}^d$ are the input gate, the output gate and the update value, respectively, and $f_{vk}$ is the forget gate associated with $k$-th child of $v$. The symbol $\sigma$ denotes the logistic sigmoid function and $\odot$ denotes the elementwise product. 

\begin{figure}
\begin{subfigure}[b]{0.24\textwidth}
        \centering
        \scalebox{0.6}{
        \begin{tikzpicture}[edge from parent/.style={draw,<-,>=triangle 45}, every node/.style = {minimum size = 0.5cm}, execute at begin node=$, execute at end node=$,node distance = 0.5cm,level 1/.style={level distance=2.5cm},level 2/.style={level distance=1.5cm, sibling distance=1cm}, level 3/.style={level distance=1.3cm},level 4/.style={level distance=1.3cm}]
        \footnotesize
		\pgfmathsetmacro{\cubex}{1}
		\pgfmathsetmacro{\cubey}{1}
		\pgfmathsetmacro{\cubez}{1}
		\pgfmathsetmacro{\corex}{0.9}
		\pgfmathsetmacro{\corey}{0.7}
		\node [minimum size=0cm] (qu) {h \in \mathbb{R}^c}
		child { 
					node (G){\scalebox{2}{$+$}}
					child{
							node [minimum size=0cm] (u_1) {
		    						\begin{tikzpicture}[every edge quotes/.append style={auto, text=blue}]
			    						\draw [draw=black, every edge/.append style={draw=black, densely dashed, opacity=.5}, fill=yellow!40]
			    						(0,0,0) coordinate (o) -- ++(-\corex,0,0) coordinate (a) -- ++(0,-\corey,0) coordinate (b) -- ++(\corex,0,0) coordinate (c) -- cycle;
			    						\node at (-0.45,-0.35) {U_l};
		    						\end{tikzpicture}
	    						}
	    						child{ node {h_l \in \mathbb{R}^c} 
	    						    edge from parent node[pos=0.7,left,draw=none] {c}
	    						}
						edge from parent node[left,draw=none] {c}  
					}
					child{
						node {} edge from parent[draw=none]
						child{
							node [minimum size=0cm] {} edge from parent[draw=none]
						}
					}
					child{
							node [minimum size=0cm] (u_1) {
		    						\begin{tikzpicture}[every edge quotes/.append style={auto, text=blue}]
			    						\draw [draw=black, every edge/.append style={draw=black, densely dashed, opacity=.5}, fill=green!40]
			    						(0,0,0) coordinate (o) -- ++(-\corex,0,0) coordinate (a) -- ++(0,-\corey,0) coordinate (b) -- ++(\corex,0,0) coordinate (c) -- cycle;
			    						\node at (-0.45,-0.35) {U_r};
		    						\end{tikzpicture}
	    						}
	    						child{ node {h_r \in \mathbb{R}^c} 
	    						    edge from parent node[pos=0.7,right,draw=none] {c}
	    						}
						edge from parent node[right,draw=none] {c}  
			    }
			edge from parent node[pos=0.7,left,draw=none] {c} 
		};
    \end{tikzpicture}
    }
\caption{Binary LSTM}\label{fig:bin_sumLSTM}\end{subfigure}
\rulesep
\begin{subfigure}[b]{0.24\textwidth}
    \centering
         \scalebox{0.6}{
        \begin{tikzpicture}[edge from parent/.style={draw,<-,>=triangle 45}, every node/.style = {minimum size = 0.5cm}, execute at begin node=$, execute at end node=$,node distance = 0.5cm,level 1/.style={level distance=1.3cm},level 2/.style={level distance=1.3cm}, level 3/.style={level distance=1.5cm, sibling distance=1cm},level 4/.style={level distance=1.3cm}]
        \footnotesize
		\pgfmathsetmacro{\cubex}{1}
		\pgfmathsetmacro{\cubey}{1}
		\pgfmathsetmacro{\cubez}{1}
		\pgfmathsetmacro{\corex}{0.9}
		\pgfmathsetmacro{\corey}{0.7}
		\node [minimum size=0cm] (qu) {h \in \mathbb{R}^c}
		child { 
				node[] (q) {
					\begin{tikzpicture}[every edge quotes/.append style={auto, text=blue}]
						\draw [draw=black, every edge/.append style={draw=black, densely dashed, opacity=.5}, fill=blue!40] (0,0,0) coordinate (o) -- ++(-\corex,0,0) coordinate (a) -- ++(0,-\corey,0) coordinate (b) -- ++(\corex,0,0) coordinate (c) -- cycle;
						\node at (-0.45,-0.35) {Q};
					\end{tikzpicture}
				}
				child {  
					node (G){\scalebox{2}{$\odot$}}
					child{
							node [minimum size=0cm] (u_1) {
		    						\begin{tikzpicture}[every edge quotes/.append style={auto, text=blue}]
			    						\draw [draw=black, every edge/.append style={draw=black, densely dashed, opacity=.5}, fill=yellow!40]
			    						(0,0,0) coordinate (o) -- ++(-\corex,0,0) coordinate (a) -- ++(0,-\corey,0) coordinate (b) -- ++(\corex,0,0) coordinate (c) -- cycle;
			    						\node at (-0.45,-0.35) {U_l};
		    						\end{tikzpicture}
	    						}
	    						child{ node {h_l \in \mathbb{R}^c} 
	    						    edge from parent node[pos=0.7,left,draw=none] {c}
	    						}
						edge from parent node[left,draw=none] {r}  
					}
					child{
						node {} edge from parent[draw=none]
						child{
							node [minimum size=0cm] {} edge from parent[draw=none]
						}
					}
					child{
							node [minimum size=0cm] (u_1) {
		    						\begin{tikzpicture}[every edge quotes/.append style={auto, text=blue}]
			    						\draw [draw=black, every edge/.append style={draw=black, densely dashed, opacity=.5}, fill=green!40]
			    						(0,0,0) coordinate (o) -- ++(-\corex,0,0) coordinate (a) -- ++(0,-\corey,0) coordinate (b) -- ++(\corex,0,0) coordinate (c) -- cycle;
			    						\node at (-0.45,-0.35) {U_r};
		    						\end{tikzpicture}
	    						}
	    						child{ node {h_r \in \mathbb{R}^c} 
	    						    edge from parent node[pos=0.7,right,draw=none] {c}
	    						}
						edge from parent node[right,draw=none] {r}  
					}
				    edge from parent node[pos=0.7,left,draw=none] {r} 
			    }
			edge from parent node[pos=0.7,left,draw=none] {c} 
		};
    \end{tikzpicture}}
\caption{Binary CP-LSTM}\label{fig:bin_canonicalLSTM}\end{subfigure}
\rulesep
\rulesep
\begin{subfigure}[b]{0.24\textwidth}
        \centering
         \scalebox{0.6}{
        \begin{tikzpicture}[edge from parent/.style={draw,<-,>=triangle 45}, every node/.style = {minimum size = 0.5cm}, execute at begin node=$, execute at end node=$,node distance = 0.5cm,level 1/.style={level distance=2.5cm},level 2/.style={level distance=1.5cm, sibling distance=1cm}, level 3/.style={level distance=1.3cm},level 4/.style={level distance=1.3cm}]
        \footnotesize
		\pgfmathsetmacro{\cubex}{1}
		\pgfmathsetmacro{\cubey}{1}
		\pgfmathsetmacro{\cubez}{1}
		\pgfmathsetmacro{\corex}{0.9}
		\pgfmathsetmacro{\corey}{0.7}
		\node [minimum size=0cm] (qu) {h \in \mathbb{R}^c}
		child { 
					node (G){\scalebox{2}{$+$}}
					child{
							node [minimum size=0cm] (u_1) {
		    						\begin{tikzpicture}[every edge quotes/.append style={auto, text=blue}]
			    						\draw [draw=black, every edge/.append style={draw=black, densely dashed, opacity=.5}, fill=red!40]
			    						(0,0,0) coordinate (o) -- ++(-\corex,0,0) coordinate (a) -- ++(0,-\corey,0) coordinate (b) -- ++(\corex,0,0) coordinate (c) -- cycle;
			    						\node at (-0.45,-0.35) {U};
		    						\end{tikzpicture}
	    						}
	    						child{ node {h_k \in \mathbb{R}^c} 
	    						    edge from parent node[pos=0.7,left,draw=none] {c}
	    						}
						edge from parent node[left,draw=none] {c}  
					}
					child{
						node {\dots} edge from parent[draw=none]
						child{
							node [minimum size=0cm] {} edge from parent[draw=none]
						}
					}
					child{
							node [minimum size=0cm] (u_1) {
		    						\begin{tikzpicture}[every edge quotes/.append style={auto, text=blue}]
			    						\draw [draw=black, every edge/.append style={draw=black, densely dashed, opacity=.5}, fill=red!40]
			    						(0,0,0) coordinate (o) -- ++(-\corex,0,0) coordinate (a) -- ++(0,-\corey,0) coordinate (b) -- ++(\corex,0,0) coordinate (c) -- cycle;
			    						\node at (-0.45,-0.35) {U};
		    						\end{tikzpicture}
	    						}
	    						child{ node {h_{k'} \in \mathbb{R}^c} 
	    						    edge from parent node[pos=0.7,right,draw=none] {c}
	    						}
						edge from parent node[right,draw=none] {c}  
			    }
			edge from parent node[pos=0.7,left,draw=none] {c} 
		};
    \end{tikzpicture}}
\caption{Child-Sum LSTM}\label{fig:inv_sumLSTM}
\end{subfigure}
\rulesep
\begin{subfigure}[b]{0.24\textwidth}
    \centering
        \scalebox{0.6}{
        \begin{tikzpicture}[edge from parent/.style={draw,<-,>=triangle 45}, every node/.style = {minimum size = 0.5cm}, execute at begin node=$, execute at end node=$,node distance = 0.5cm,level 1/.style={level distance=1.3cm},level 2/.style={level distance=1.3cm}, level 3/.style={level distance=1.5cm, sibling distance=1cm},level 4/.style={level distance=1.3cm}]
        \footnotesize
		\pgfmathsetmacro{\cubex}{1}
		\pgfmathsetmacro{\cubey}{1}
		\pgfmathsetmacro{\cubez}{1}
		\pgfmathsetmacro{\corex}{0.9}
		\pgfmathsetmacro{\corey}{0.7}
		\node [minimum size=0cm] (qu) {h \in \mathbb{R}^c}
		child { 
				node[] (q) {
					\begin{tikzpicture}[every edge quotes/.append style={auto, text=blue}]
						\draw [draw=black, every edge/.append style={draw=black, densely dashed, opacity=.5}, fill=blue!40] (0,0,0) coordinate (o) -- ++(-\corex,0,0) coordinate (a) -- ++(0,-\corey,0) coordinate (b) -- ++(\corex,0,0) coordinate (c) -- cycle;
						\node at (-0.45,-0.35) {Q};
					\end{tikzpicture}
				}
				child {  
					node (G){\scalebox{2}{$\odot$}}
					child{
							node [minimum size=0cm] (u_1) {
		    						\begin{tikzpicture}[every edge quotes/.append style={auto, text=blue}]
			    						\draw [draw=black, every edge/.append style={draw=black, densely dashed, opacity=.5}, fill=red!40]
			    						(0,0,0) coordinate (o) -- ++(-\corex,0,0) coordinate (a) -- ++(0,-\corey,0) coordinate (b) -- ++(\corex,0,0) coordinate (c) -- cycle;
			    						\node at (-0.45,-0.35) {U};
		    						\end{tikzpicture}
	    						}
	    						child{ node {h_k \in \mathbb{R}^c} 
	    						    edge from parent node[pos=0.7,left,draw=none] {c}
	    						}
						edge from parent node[left,draw=none] {r}  
					}
					child{
						node {\dots} edge from parent[draw=none]
						child{
							node [minimum size=0cm] {} edge from parent[draw=none]
						}
					}
					child{
							node [minimum size=0cm] (u_1) {
		    						\begin{tikzpicture}[every edge quotes/.append style={auto, text=blue}]
			    						\draw [draw=black, every edge/.append style={draw=black, densely dashed, opacity=.5}, fill=red!40]
			    						(0,0,0) coordinate (o) -- ++(-\corex,0,0) coordinate (a) -- ++(0,-\corey,0) coordinate (b) -- ++(\corex,0,0) coordinate (c) -- cycle;
			    						\node at (-0.45,-0.35) {U};
		    						\end{tikzpicture}
	    						}
	    						child{ node {h_{k'} \in \mathbb{R}^c} 
	    						    edge from parent node[pos=0.7,right,draw=none] {c}
	    						}
						edge from parent node[right,draw=none] {r}  
					}
				    edge from parent node[pos=0.7,left,draw=none] {r} 
			    }
			edge from parent node[pos=0.7,left,draw=none] {c} 
		};
    \end{tikzpicture}}
\caption{Invariant CP-Tree-LSTM}\label{fig:inv_canonicalLSTM}
\end{subfigure}

\caption{Graphical illustration of composition functions; for the sake of clarity, we removed the bias.}
\label{fig:tensorLSTM}
\end{figure}
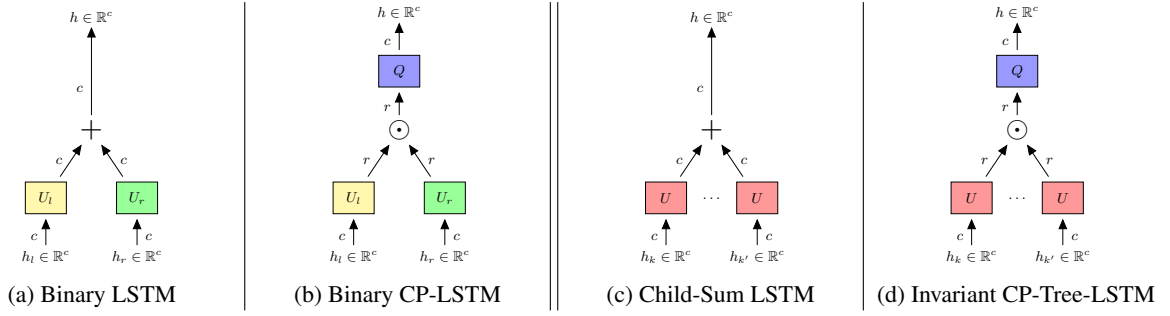

We apply this model on binary constituency tree. Hence, the input tree has labels (i.e.\ the words) attached only on leaf nodes. In particular, we consider $x_v$ as the vector representation of a word in the sentence. Hence, internal nodes do not have any input labels (see Fig.\ \ref{fig:bin_sumLSTM}). In the rest of the paper, we refer to this model as Binary Sum-LSTM.

\paragraph{Child-Sum Tree-LSTM} \cite{Tai2015} assumes that there is no order among child nodes. Let $v$ a generic node, we indicate with $\text{Ch}(v)$ the set of its child nodes; hence, its hidden state $h_v$ is computed as:
\begin{equation}
    \begin{alignedat}{2}
        t_v &= \sigma\left( W^tx_v + \sum_{k \in \text{Ch}(v)}U^t h_{k} + b^t\right) \quad t \in \{i,o,u\}, \quad & \quad c_v &= i_v \odot u_v + \sum_{k \in \text{Ch}(v)} f_{vk} \odot c_{vk},\\
        f_{vk} &= \sigma\left( W^fx_v + U^f h_{vk} + b^f\right) \quad k \in \text{Ch}(v), \quad & \quad h_v &= o_v \odot \text{tanh}(c_v).
    \end{alignedat}
    \label{eq:childsumLSTM}
\end{equation}

When comparing Eq.\ \eqref{eq:childsumLSTM} and Eq.\ \eqref{eq:laryLSTM}, we observe that the Child-Sum Tree-LSTM can be derived from the $L$-ary Tree-LSTM by imposing weight sharing across children positions. In fact, in Eq.\ \eqref{eq:childsumLSTM}, the parameter $U$ is the same for each child, while in in Eq.\ \eqref{eq:laryLSTM} the subscript $j$ of the parameter $U_j$ denotes that the weights are positional dependent.

In \newcite{Tai2015}, this model has been applied only on dependency trees. However, in this work, we use it on non-binary constituency trees. Again, we consider $x_v$ as the vector representation of a word in the sentence and therefore are attached only on leaf nodes (see Fig.\ \ref{fig:inv_sumLSTM}). In the remainder of the paper, we refer this model with the name Child-Sum LSTM.

\subsection{TreeNet}\label{sec:treeNET}
TreeNet \cite{Cheng2018} has been introduced with the aim of learning from unconstrained tree-structured data (i.e.\ trees where out-degree not fixed). The idea is to process child nodes as a sequence and links only the rightmost child node to the parent node. Hence, each node compose the information of its left sibling and its rightmost child:
\begin{equation}
    \begin{alignedat}{2}
        o_v &= \sigma\left(U^o_s h_{vs} + U^o_c h_{vc} + b^o\right), \quad & \quad c_v &= f_{vs} \odot c_{vs} + f_{vc} \odot c_{vc},\\
        f_{vk} &= \sigma\left(U^f_{ks} h_{vs} + U^f_{kc} h_{vc} + b^f_{k}\right), \quad k \in \{s,c\},  \quad & \quad h_v &= o_v \odot \text{tanh}(c_v),
    \end{alignedat}
    \label{eq:treeNet}
\end{equation}
where $\{h_{vs}, c_{vs}\} \in \mathbb{R}^d$ are hidden state and memory cell of left sibling of $v$, while $\{h_{vc}, c_{vc}\} \in \mathbb{R}^d$ are hidden state and memory cell of the rightmost child node of $v$.

If the node $v$ is a leaf, its hidden state depends solely on the input label $x_v$:
\begin{equation}
    \begin{alignedat}{3}
        i_v &= \sigma\left(W^i x_v + \hat{b}^i\right), \quad & \quad o_v &= \sigma\left(W^o x_v + \hat{b}^o\right), \quad & \quad u_v &= \sigma\left(W^u x_v + \hat{b}^u\right),\\
        c_v &= i_v \odot u_v, \quad & \quad h_v &= o_v \odot \text{tanh}(c_v).
    \end{alignedat}
    \label{eq:treeNet_leaf}
\end{equation}

Comparing Eq.\ \eqref{eq:treeNet} and Eq.\ \eqref{eq:laryLSTM}, we can observe that TreeNet cell is a binary Tree-LSTM cell without the input gate and the update value. In fact, in both cells, all the gates are computed composing two constituents. The TreeNet define the constituents of a node as its left sibling and its rightmost child, while the binary Tree-LSTM used directly its left child node and right child node. Hence, we argue that the difference between the two lies solely on how the tree is binarised, rather than on how the tree is processed. In Fig.\ \ref{fig:tree_net_tree}, we show an example on how a constituency tree is binarised according to the TreeNet; the constituent \emph{ADJP} of the original constituency tree (see Fig.\ \ref{fig:orig_tree}) is composed of three words: "\emph{Effective}", "\emph{but}", "\emph{too-tepid}". The TreeNet breaks this ternary relation processing one words at a time; as we can see in Fig.\ \ref{fig:tree_net_tree}, the node which has the first word "\emph{Effective}" is combined with a bottom node since it does not have a left sibling. The result is then fused with the word "\emph{but}". Finally, also the word "\emph{too-tepid}" is combined with the result of the previous composition, obtaining the encoding of the constituent \emph{ADJP}. Hence, the original ternary relation is broken into a sequence of three binary relations (one for each word) each of them combines the composition of previous words with the new word.

\section{Canonical Tree-LSTM}
This section aims to introduce new Tree-LSTM models that extend the recurrent approach to tensor-based processing tailored for constituency trees. 
These two models rely on Tensor Tree-LSTM \cite{Castellana2020} and canonical tensor decomposition. We focus on such decomposition since it can be combined with a weight sharing constraint, developing a composition function which can exploit non-binary constituency trees without increasing the model parameters number.

Since these models are applied on constituency trees, we define them only on the nodes which apply composition functions (i.e.\ the internal nodes). The input label $x_v$, which is attached only on leaf nodes, is processed using the same computation described in Eq.\ \ref{eq:treeNet_leaf}.


\subsection{Canonical Decomposition}
The canonical decomposition (usually denoted by CP) factorises a tensor into a sum of component rank-one tensors, i.e. tensors that are obtained by the outer product of vectors. For example, let $\Tensor{T}\in \mathbb{R}^{d_1 \times d_2 \times \dots \times d_L}$ a general $L$-th order tensor, the CP decomposition is defined by \cite{Kolda2009}:
\begin{equation}
    \Tensor{T}(j_1, \dots, j_L) = \sum_{i=1}^r U_1(j_1,i) U_2(j_2,i) \dots U_L(j_L,i),
\end{equation}
where brackets are used to denote entries of vectors, matrices and tensors (e.g.\ $W(i,j)$ denotes the entry in the $i$-th row and $j$-th column of $W$). $U_1 \in \mathbb{R}^{d_1 \times r}, \dots, U_L  \in \mathbb{R}^{d_L \times r}$ are the factor matrices of the decomposition. Each factor matrix contains $r$ vectors, which are the basis of the rank-one tensors. The value of $r$ indicates the number of such tensors which are summed to obtain the original tensor \Tensor{T} and it is denoted as tensor rank.

Following \newcite{Castellana2020}, we apply the decomposition on a tensor which represents a multi-affine map. Note that in this case the tensor that should be decomposed in not known, since it is the parameter of the recursive model and it is learned from data. Instead, we assume that such tensor is already decomposed, making the decomposition factors the new recursive model parameters. Let $\phi^\Tensor{T}: \mathbb{R}^{d_1} \times \dots \times \mathbb{R}^{d_1} \rightarrow \mathbb{R}^K$ a multi-affine map which parameter is the tensor  $\Tensor{T} \in \mathbb{R}^{(d_1+1) \times \dots \times (d_L+1) \times K}$, applying the CP decomposition on \Tensor{T} we obtain:
\begin{multline}
    a(k) = \phi^\Tensor{T}(a_{1},\dots,a_{L})(k) =\sum_{j_1=1}^{d_1+1}\dots\sum_{j_L=1}^{d_L+1}\Tensor{T}(j_1, \dots,j_L, k) \bar{a}_1(j_1)\dots \bar{a}_L(j_L) =\\
    = \sum_{j_1=1}^{d_1+1}\dots\sum_{j_L=1}^{d_L+1}\sum_{i=1}^r U_1(j_1,i) \dots U_L(j_L,i)Q(k,i) \bar{a}_1(j_1)\dots \bar{a}_L(j_L)=\\
    = \sum_{i=1}^r  Q(k,i) \left(\sum_{j_1=1}^{d_1+1}U_1(j_1,i)\bar{a}_1(j_1)\right) \dots \left(\sum_{j_L=1}^{d_L+1}U_L(j_L,i)\bar{a}_L(j_L)\right),
    \label{eq:canonical_comp_fun}
\end{multline}
where $\bar{a} = [a; 1]$ denotes the homogeneous coordinate of vector $a$. From the equation, we can observe that the CP decomposition define a multi-affine map which applies each factor matrix to the corresponding input vector (i.e. the one on the same mode), obtaining a vector $e_j \in \mathbb{R}^r$, for each mode $j \in \{1, \dots,L\}$. Then, these vectors are element-wise multiplied and the result mapped to the output space $\mathbb{R}^K$ thanks to matrix $Q$, i.e. the factor matrices on the output (the last) mode.

\subsection{Binary Canonical Tree-LSTM}

The Binary Tensor Tree-LSTM \cite{Castellana2020} computes the hidden state of an intern node $v$ by:
\begin{equation}
    \begin{alignedat}{2}
        i_v &= \sigma\left(\phi^{\Tensor{I}}(h_{vl}, h_{vr})\right), \quad & \quad f_{vk} &= \sigma\left(\phi^{\Tensor{F}_k}(h_{vl}, h_{vr})\right) \text{ with } k\in \{l,r\},\\
        o_v &= \sigma\left(\phi^{\Tensor{O}}(h_{vl}, h_{vr})\right), \quad  & \quad c_v &= i_v \odot u_v + f_{vl} \odot c_{vl} + f_{vr} \odot c_{vr},\\
        u_v &= \sigma\left(\phi^{\Tensor{U}}(h_{vl}, h_{vr})\right), \quad  & \quad h_v &= o_v \odot \text{tanh}(c_v).
    \end{alignedat}
    \label{eq:bin_tensor_LSTM}
\end{equation}
All the terms (except hidden state and memory cell) are computed by applying a multi-affine map $\phi^{\Tensor{T}}(\cdot)$ on the left and right child hidden states, i.e. $h_{vl}$ and $h_{vr}$, respectively. The superscript \Tensor{T} indicates the parameters of such multi-affine map; hence, the parameters $\{\Tensor{I}, \Tensor{O}, \Tensor{U}, \Tensor{F}_l, \Tensor{F}_r\}$ define the LSTM cell computation. 


The Binary Canonical Tree-LSTM (Binary CP-LSTM) exploits the canonical decomposition on tensor parameters (see Fig.\ \ref{fig:bin_canonicalLSTM}): 
\begin{equation}
    \begin{alignedat}{4}
        e_l &= U^t_l h_{vl} + b^t_l, \quad & \quad e_r &= U^t_l h_{vr} + b^t_r,\quad & \quad e &=a_l \odot a_r, \quad & \quad \phi^\Tensor{T}(h_{vl}, h_{vr}) &= Q^t e + b^t,
    \end{alignedat}
    \label{eq:canonicalLSTM}
\end{equation}
where $\{U^t_l,  U^t_r\} \in \mathbb{R}^{d \times r}$, $Q^t \in \mathbb{R}^{r \times d}$, $\{b^t_l,b^t_r\} \in \mathbb{R}^r$, $b \in \mathbb{R}^d$ define the decomposition's factor matrices (the bias is made explicit) and the value $r$ is the decomposition rank. We use the superscript $t \in \{i,o,u,f_l,f_r\}$ to indicate different parameters of each multi-affine maps. The number of parameters required by the Binary CP-LSTM is $O(dr)$.

\subsection{Invariant Canonical Tree-LSTM}
The Invariant Canonical Tree-LSTM (Invariant CP-LSTM) allows to encode information from constituency trees which are not binarised. Let $v$ an internal node, its hidden state it is computed by:
\begin{equation}
    \begin{alignedat}{2}
        i_v &= \sigma\left(\,\phi^{\Tensor{I}}(\text{Ch}(v))\,\right), \quad & \quad f_{vk} &= \sigma\left(U^fh_{vk} + b^f\right) \text{ with } k\in \text{Ch}(v),\\
        o_v &= \sigma\left(\,\phi^{\Tensor{O}}(\text{Ch}(v))\,\right), \quad  & \quad c_v &= i_v \odot u_v + \sum_{k \in \text{Ch}(v)} f_{vk} \odot c_{vk},\\
        u_v &= \sigma\left(\,\phi^{\Tensor{U}}(\text{Ch}(v))\,\right), \quad  & \quad h_v &= o_v \odot \text{tanh}(c_v).
    \end{alignedat}
    \label{eq:stat_tensor_LSTM}
\end{equation}
As in Eq.\ \eqref{eq:childsumLSTM}, we indicate with $\text{Ch}(v)$ the set of child nodes of $v$. $\{\Tensor{I}, \Tensor{O}, \Tensor{U}\}$ are the parameters of the multi-affine maps to compute the input gate, output gate and the update value respectively.


Again, we exploit the canonical decomposition on parameter tensors to define the Invariant CP-LSTM. Moreover, we impose the weight sharing among input factor matrices, obtaining (see Fig.\ \ref{fig:inv_canonicalLSTM}):
\begin{equation}
    \begin{alignedat}{3}
        e_k &= U^t h_k + b^t \text{ with } k\in \text{Ch}(v), \quad & \quad e &= \bigodot_{k \in \text{Ch}(v)} e_k, \quad & \quad \phi^\Tensor{T}(\text{Ch}(v)) &=  Q^t e + q^t.
    \end{alignedat}
\end{equation}
where $U^t \in \mathbb{R}^{d \times r}, b^t \in \mathbb{R}^r $ is the factor matrix shared on all input modes and $Q^t \in \mathbb{R}^{r \times d}$, $q^t \in \mathbb{R}^d$ is the factor matrix on the last mode; the value $r$ is the decomposition rank. Hence, the number of parameters required by Invariant CP-LSTM is still $O(dr)$. Thanks to the weight sharing, we are able to process non-binary trees without adding new parameters. As we highlighted in Sec. \ref{sec:treeLSTM}, the weight sharing makes the model \emph{invariant} to child nodes order, thus its name is Invariant CP-LSTM.


\section{Experiments}\label{sec:experiment}
We test the models introduced in the previous section on two tasks: sentence classification and semantic textual similarity.


\paragraph{Sentence Classification.} The goal of this task is to predict the class of the given input sentence. Hence, we use a Tree-LSTM to encode the constituency tree of the input sentence in a succinct representation ($h_{\text{root}}$) and then we feed it to a single-layer Neural Network to predict the class $y \in \{1\dots m\}$:
\begin{equation}
    \begin{alignedat}{2}
    s &= \text{ReLU}\left(W' h_{\text{root}} + b'\right) \quad & \quad p(y) &= \text{softmax}\left(W'' s + b''\right),
    \end{alignedat}
\end{equation}
where $s \in \mathbb{R}^s$ is the hidden representation of the classifier and $W' \in \mathbb{R}^{d\times s}$, $b \in \mathbb{R}^s, W'' \in \mathbb{R}^{s\times m}, b'' \in \mathbb{R}^m$ are the classifiers parameters. Following \newcite{Tai2015}, we use dropout \cite{Srivastava2014} with rate 0.5 on both $h_{root}$ and $h_s$. We test our models on three different classification datasets:
\begin{itemize}
    \item \textbf{SST-5}: Stanford Sentiment Treebank (SST) dataset \cite{Socher2013} contains sentences that are classified with a fine-grained sentiment which goes from $1$ to $5$ (very negative, negative, neutral, positive and very positive);
    \item \textbf{SST-2}: identical to SST-5, but with binary sentiment class; neutral sentences are removed and all negative (positive) sentences are collapsed in one cluster;
    \item \textbf{TREC}: TREC dataset \cite{Li2002} contains questions that are annotated with six classes which indicates a question type.
\end{itemize}

\paragraph{Semantic Textual Similarity.} The goal of this task is to predict the semantic similarity between two sentences. Let $a$ and $b$ the two sentences, we produce their encodings $h_a$ and $h_b$ applying a Tree-LSTM on both sentence constituency trees and taking the hidden state of the root. Then, we compute the similarity score $y_r \in [1, m]$ as in \newcite{Tai2015}:
\begin{equation}
    \begin{alignedat}{3}
    s &= \sigma\left(W^+|h_a - h_b| + W^\times (h_a \odot h_b) + b\right), \quad & \quad p_r &= \text{softmax}\left(W^r s + b^r\right), \quad & \quad y_r &= r^\top p_r \quad & \quad &
    \end{alignedat}
    \label{eq:semantic_score}
\end{equation}
where $s \in \mathbb{R}^s$ is the hidden representation of the classifier, $\{W^+, W^\times \} \in \mathbb{R}^{d\times s}$, $b \in \mathbb{R}^s, W^r \in \mathbb{R}^{s\times m}, b^r \in \mathbb{R}^m$ are the classifiers parameters and $r^\top = [1, 2, \dots, m]$.

It also common to represent the semantic similarity attaching an entailment class to each pair of sentences. In this case, we predict the entailment class $y_e= \argmax(p_e)$ starting from the distribution $p_e = \text{softmax}\left(W^e s + b^e\right)$; $s$ is computed as in Eq.\ \eqref{eq:semantic_score}. We test our models on two different datasets:
\begin{itemize}
    \item \textbf{SICK-R}: Sentences Involving Compositional Knowledge (SICK) dataset \cite{Marelli2015}, contains sentence pairs annotated with a relatedness score between $1$ and $5$;
    \item \textbf{SICK-E}: identical to SICK-R, but with entailment classes instead of relatedness scores; the entailment class indicates whether one sentence entails or contradicts the other (neutral, entailment, contradiction).
\end{itemize}

\subsection{Implementation and training details}
We have implemented all the models using PyTorch \cite{Paszke2019} and Deep Graph Library \cite{Wang2019}. The source code to reproduce the model and the experiments is released here.\footnote{https://github.com/danielecastellana22/tensor-tree-nn} Constituency trees are built using the PCFG constituency parser of the Stanford Core NLP \cite{Manning2014}. Also, we binarise them computing the Chomsky Normal Form available in the Natural Language Tool Kit \cite{bird2009natural}. To facilitate the learning, we collapse all unary relations. In each task, we perform a grid search to find the best hyper-parameters configuration (see Appendix \ref{sec:exp_details} for further details). For the training, we use AdaDelta \cite{Zeiler2012} algorithm and Adam \cite{Kingma2015} optimiser. In all experiments, we initialised our word representations using $300$-dimensional Glove vectors \cite{Pennington2014}. We fine-tune the words representation only on the SST dataset.

\section{Results}

\begin{figure}
    \centering
    \begin{subfigure}[t]{0.45\textwidth}
    \centering
    \includegraphics[width=0.9\textwidth]{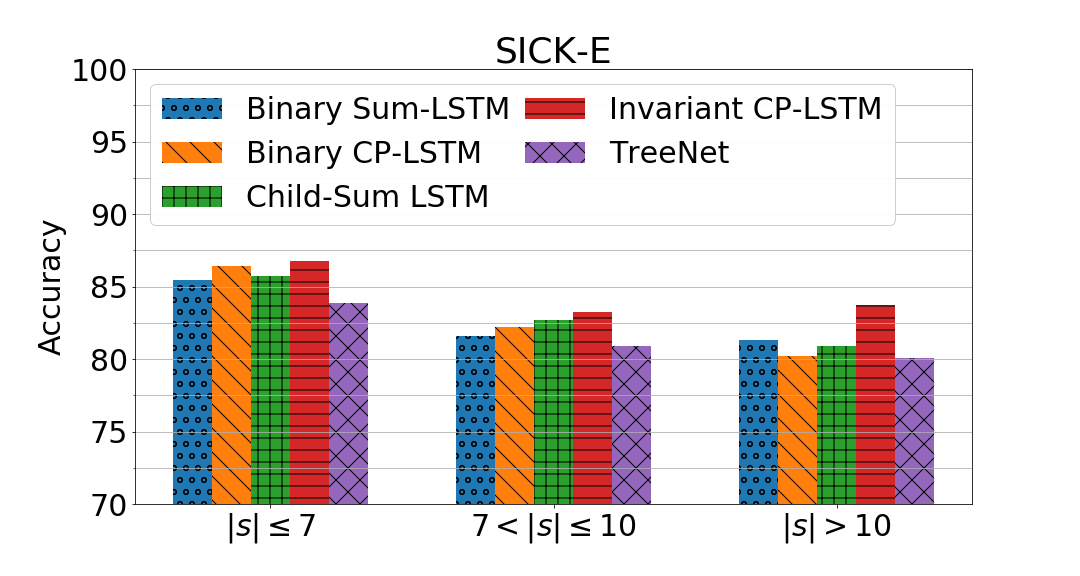}
    \caption{Test accuracy on SICK-E w.r.t. sentences length.}
    \label{fig:sick_e_erro_wrt_size}
    \end{subfigure}
    %
    \begin{subfigure}[t]{0.45\textwidth}
    \centering
    \includegraphics[width=0.9\textwidth]{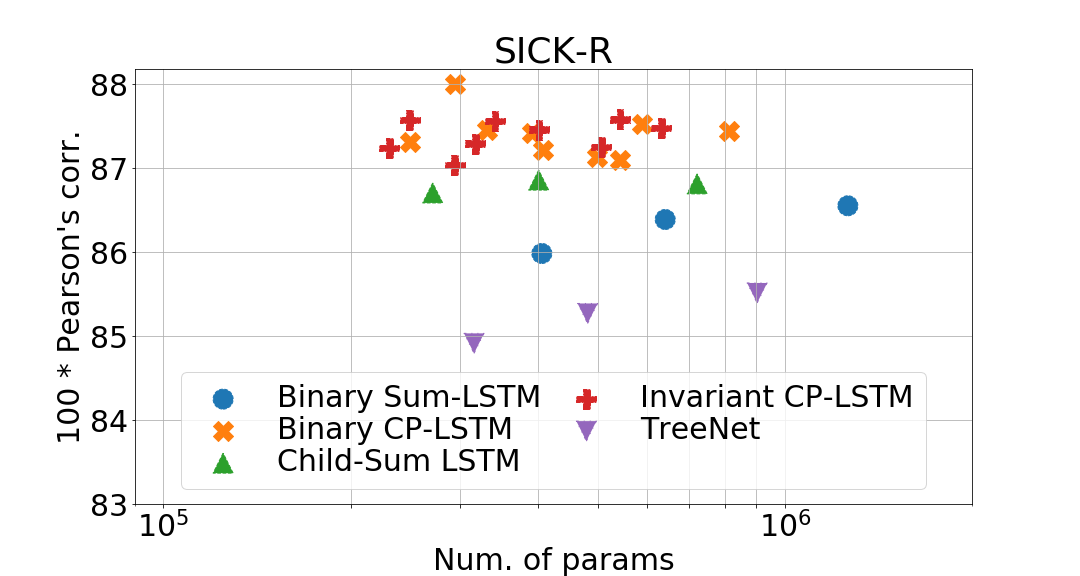}
    \caption{Validation Pearson's corr. on SICK-R w.r.t. the number of parameters.}
    \label{fig:sick_r_num_params}
    \end{subfigure}
    \caption{Analyses of the results obtained on SICK-E and SICK-R dataset.}
\end{figure}

Table \ref{tab:all_results} reports the results obtained in our experiments. Also, we report the results obtained by our implementation of the TreeNet \cite{Chen2018} which differ from the results published in the original paper on the TREC dataset. However, to be sure that there are no errors in our implementation, we run the original code published by \newcite{Cheng2018} with our experimental settings (i.e. model selection on the validation set and risk assessment on the test set) and we obtain results comparable with the one published in this table. In the next paragraphs, we analyse in details the results obtained on each dataset.

\begin{table}
    \centering
    \scriptsize
    \begin{tabular}{rccccc}
        \hline 
        & \textbf{SST-5}   & \textbf{SST-2}   & \textbf{SICK-E}   & \textbf{SICK-R}   & \textbf{TREC}   \\
        \hline  
        \textbf{Binary Sum-LSTM}*       & 51.5 (0.7)       & 87.9 (0.2)       & 82.3 (0.5)        & 84.3 (0.7)        & \textbf{92.3 }(0.8)      \\
        \textbf{Binary CP-LSTM} (our) & 50.3 (0.8)       & 88.0 (0.3)       & 82.4 (0.3)        & 85.9 (0.1)        & 90.4 (0.2)      \\
        \textbf{Binary LSTM} \cite{Tai2015} & 51.00 (0.5) & 88.0 (0.3) & - & 85.8 (0.4) & - \\
         \hline
         \hline
         \textbf{Child-Sum LSTM}*       & 49.4 (0.6)       & 85.5 (0.8)       & 82.6 (0.4)        & 84.9 (0.2)        & 91.9 (1.0)      \\
         \textbf{Invariant CP-LSTM} (our) & 48.3 (0.8)       & 85.3 (0.3)       & \textbf{84.2} (0.4)        & \textbf{86.4} (0.1)        & 90.0 (0.7)      \\
         \textbf{TreeNet}*             & 48.4 (1.5)       & 87.0 (0.5)       & 81.2 (0.3)        & 84.5 (0.3)        & 91.3 (1.1)      \\
        \hline
    \end{tabular}
    \caption{Results obtained on different task All the values are accuracy except SICK-R, whose score is Pearson's correlation multiplied by 100. The superscript * indicates we re-implement the model.}
    
    \label{tab:all_results}
\end{table}

\paragraph{SST.} The results obtained on the SST dataset (both with fine-grained and binary labels) do not show any improvements using non-binary constituency trees. However, the comparison is unfair since the original dataset provides labels on internal nodes of binary constituency trees.  By removing the binarisation, it is no longer possible to leverage such information during training. As a reference, note that binary constituency trees data contains 119.413 labels, while the non-binary constituency trees data contains only 91.536 labels.

\paragraph{SICK.} The results obtained on the SICK dataset show the advantage of combining a rich representation (such as non-binary trees) and a more powerful composition function. In fact, the Invariant CP-LSTM outperforms all the other models in both the entailment (SICK-E) and relatedness (SICK-R) task. It is worth to point out that the Invariant CP-LSTM is the only model which benefits from the non-binary representation. In Fig.\ \ref{fig:sick_e_erro_wrt_size} we report the test accuracy of each model with respect to the input length (we consider the maximum between the length of each sentence in the input pairs). Observing the plot, it is clear that most of the models struggle with long sentences. The Invariant CP-LSTM model, instead, reaches an accuracy of approximately 84\%, while other models stop around 81\%. In Fig.\ \ref{fig:sick_r_num_params}, we show the validation results on SICK-R obtained by all the models against the number of parameters they require. Observing the plot, it is clear that thanks to the canonical decomposition and the weight sharing, we can build powerful composition function using the same number of parameters of sum-based functions.

\paragraph{TREC.} On the TREC dataset, the sum-based composition functions seems to be advantageous over the canonical counterpart. In fact, both on binary and non-binary trees, the sum-based Tree-LSTMs outperforms the canonical-based Tree-LSTMs. Also, the Tree-Net (which is based on summation), reaches results comparable with the Child-Sum LSTM. We argue that the summation is preferable for the question classification task, probably due to the intrinsic characteristics of question sentences. 

\subsection{Qualitative analysis on SICK-E}
In this section, we analyse in details the prediction of Child-Sum LSTM and Invariant CP-LSTM on the SICK-E dataset. In particular, we report the prediction on example \#3991 of the test set. The input pair is composed of the following sentences:
\begin{quote}
    A: \emph{The girl has red hair and eyebrows, several piercings in a ear and \textbf{a tattoo} on the back.
    }\\
    B: \emph{The girl has red hair and eyebrows, several piercings in a ear and \textbf{no tattoo} on the back.}
\end{quote}

\begin{figure}

\begin{subfigure}[b]{0.45\textwidth}
    \centering
    \small
    \scalebox{0.55}{
    \begin{tikzpicture}[edge from parent/.style={draw,<-,>=triangle 45}, grow=down, level 1/.style={level distance=1cm}, every node/.style = {minimum size = 0.6cm}]
        \node[internal,selected] {0}
             [sibling distance=3cm] 
             child{
            	node[internal] {1}
            	[sibling distance=1cm] 
            	child{
            		node[] {the}
            	}
            	child{
            		node[] {girl}
            	}
            }
            child{
            	node[internal, selected] {4}
            	[sibling distance=1cm] 
            	child{
            		node[] {has}
            	}
            	child{
            		node[internal,selected] {6}
            		[sibling distance=1.8cm] 
            		child{
            			node[internal] {7}
            			[sibling distance=1cm] 
            			child{
            				node[internal] {8}
            				child{
            					node[] {red}
            				}
            				child{
            					node[] {hair}
            				}
            			}
            			child{
            				node[] {and}
            			}
            			child{
            				node[] {eyebrows}
            			}
            		}
            		child{
            			node[] {,}
            		} 
            		child{
            			node[internal] {14}
            			[sibling distance=2cm] 
            			child{
            				node[internal] {15}
            				[sibling distance=1.2cm] 
            				child{
            					node[] {several}
            				}
            				child{
            					node[] {piercings}
            				}
            			}
            			child{
            				node[internal] {18}
            				[sibling distance=1cm] 
            				child{
            					node[] {in}
            				}
            				child{
            					node[internal] {20}
            					child{
            						node[] {a}
            					}
            					child{
            						node[] {ear}
            					}
            				}
            			}
            		}
            		child{
            			node[] {and}
            		}
            		child{
            			node[internal, selected] {24}
            			child{
            				node[internal, selected] {25}
            				[sibling distance=1cm] 
            				child{
            					node[] {\Large \textbf{?}}
            				}
            				child{
            					node[] {tattoo}
            				}
            			}
            			child{
            				node[internal] {28}
            				[sibling distance=1cm] 
            				child{
            					node[] {on}
            				}
            				child{
            					node[internal] {30}
            					child{
            						node[] {the}
            					}
            					child{
            						node[] {back}
            					}
            				}
            			}
            		}
            	}
            };
    \end{tikzpicture}}
\caption{Input constituency trees.}\label{fig:3991_const_tree}
\end{subfigure}
\rulesep
\begin{subfigure}[b]{0.50\textwidth}
        \centering
        \includegraphics[width=1\textwidth]{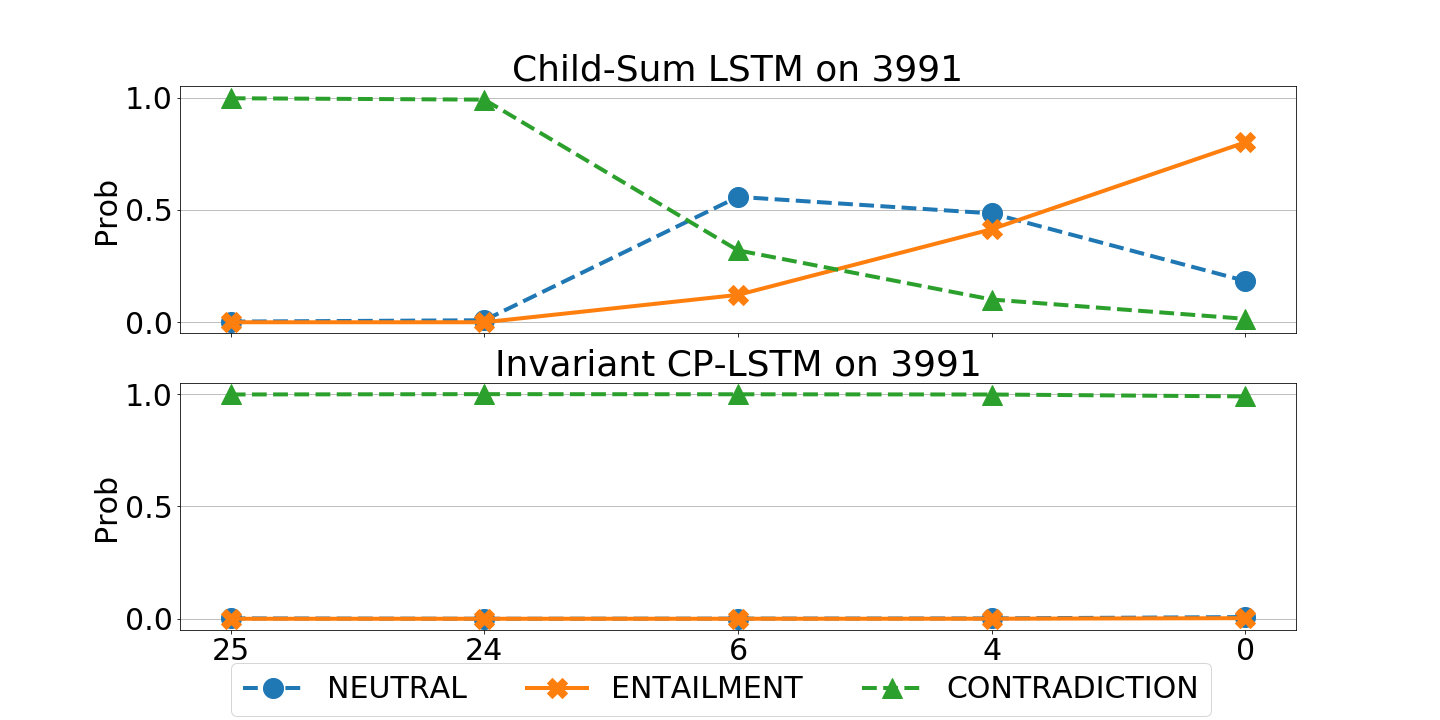}
    \caption{Model prediction on different nodes.}
    \label{fig:3991_inv_predictions}
\end{subfigure}
\caption{Comparison between Invariant CP-LSTM and Child-Sum LSTM prediction on input \#3991 taken from SICK \cite{Marelli2015} test set.}
\end{figure}
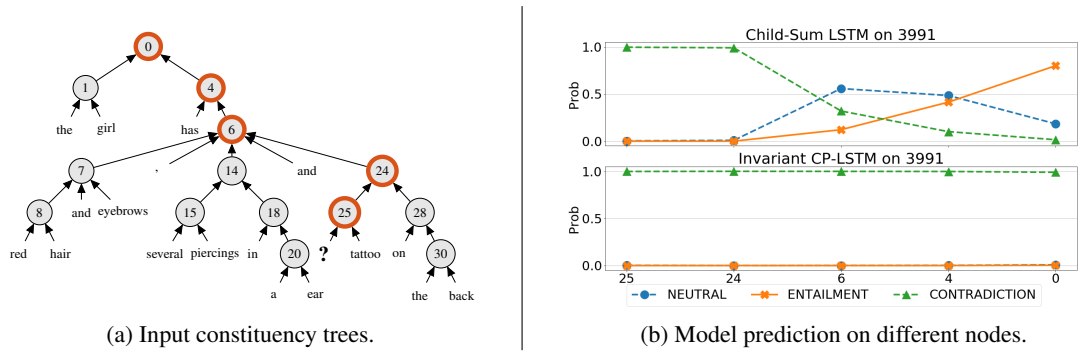

The two sentences are exactly the same, unless the sub-phrase \emph{a tattoo} in sentence A which becomes \emph{no tattoo} in sentence B. Hence, the expected output is "contradiction". In Fig.\ \ref{fig:3991_const_tree} we plot the constituency tree of the input sentence, indicating with \textbf{?} the position where the two sentences differ. Also, we highlight all the nodes that are in the path between the \textbf{?} and the root. These are the only nodes which have a different sub-tree in the two sentences. To analyse how the two models predict the final label, we study how the prediction changes going up through the structure. In Fig.\ \ref{fig:3991_inv_predictions} we report the output of the classifier fed with the hidden states pair $(h_{ai}, h_{bi})$, where $h_{ai}$ ($h_{bi}$) is the hidden state of node $i$ computed by the tree model on sentence A (B). On node 25, both models predict correctly the contradiction. However, going up through the structure, the Child-Sum Tree-LSTM changes the prediction to entailment, which will be the final output on node 0. The change of the output label start on node 6, which is the node where most of the information is aggregated; it seems that the Child-Sum Tree-LSTM performs a sort of average which soften the contribution of node 24 (the only one that instead should be taken into account). On the contrary, the Invariant CP-LSTM propagates correctly the information through the structure. Even if the model process sub-phrases which are identical, their contribution does not influence the output. In fact, observing Fig.\ \ref{fig:3991_inv_predictions}, we can notice that the class predicted by the Invariant CP Tree-LSTM is always "contradiction" in all nodes in the path between \textbf{?} and the root.

\section{Conclusion}
In this paper, we show that using non-binary constituency trees can be beneficial, especially in semantic similarity tasks. Moreover, we highlight the need of powerful composition function to exploit such a rich representation. To this end, we have introduced a new Tree-LSTM model which leverages tensor canonical decomposition and weight sharing to process non-binary trees without adding new parameters.

Such results pave the way to the definition of new tensor models which leverage suitable tensor decomposition to take advantage of non-binary constituency trees. To this end, the next step would be the application of other tensor decompositions. Among the others, the tensor train decomposition seems to be promising to define new composition functions which are sensitive to child nodes order.

Ultimately, we would like to test multiple tensor-based models on different NLP tasks, studying the relation between the bias introduced by each different tensor decomposition and the intrinsic property of the task.

\section*{Acknowledgements}
This work has been partially supported by the MIUR SIR 2014 LIST-IT project (grant n. RBSI14STDE) and by TAILOR, a project funded by EU Horizon 2020 programme under GA No 952215.

%

\bibliographystyle{coling}
\bibliography{references, new}

\clearpage

\appendix

\section{Appendix A. Experiments Details} \label{sec:exp_details}
For each model, we select the best hyper-parameters through a grid search (see Table \ref{tab:grid_search}). The hyper-parameters validated are: the batch size $bs$, the hidden representation of the Tree-LSTM $d$ and the hidden representation of the classifier $s$. The models which use the canonical decomposition have also  the rank $r$ as hyper-parameter. Moreover, we train all the models using the AdaDelta algorithm, expect for the TreeNet on SICK-E, SICK-R and TREC dataset where we use Adam optimiser. When we use Adam, we validate the learning rate $lr$ and we fix the batch size to 25.

\begin{table}[ht]
\footnotesize
\centering
\begin{tabular}{c|r|c|c|c|c}
    & \textbf{Model} & $bs$ or $lr$ & $d$ & $r$ & $s$ \\
    \hline
    \multirow{5}{*}{\rotatebox[origin=c]{90}{\textbf{SST}}} & \textbf{Binary Sum-LSTM} & $[5, 10, 25]$ & $[100, 200, 300]$ & - & $[0, 500, 1000]$\\
    & \textbf{Binary CP-LSTM} & $[5, 10, 25]$ &  $[100, 200, 300]$ & $[50, 100, 150]$ & $[0, 500, 1000]$\\
    & \textbf{Child-Sum LSTM} & $[5, 10, 25]$ & $[100, 200, 300]$ & - & $[0, 500, 1000]$\\
    & \textbf{Invariant CP-LSTM} & $[5, 10, 25]$ &  $[100, 200, 300]$ & $[50, 100, 150]$ & $[0, 500, 1000]$\\
    & \textbf{TreeNet} & $[5, 10, 25]$ & $[100, 200, 300]$ & - & $[0, 500, 1000]$ \\
    \hline
    \hline
    \multirow{5}{*}{\rotatebox[origin=c]{90}{\textbf{SICK}}} & \textbf{Binary Sum-LSTM} & $[10, 25, 40]$ & $[150, 200, 300]$ & - &  $[50, 100, 200]$\\
	& \textbf{Binary CP-LSTM} & $[10, 25, 40]$ & $[150, 200, 300]$ &  $[30, 50, 100]$ &  $[50, 100, 200]$\\
    & \textbf{Child-Sum LSTM} & $[10, 25, 40]$ & $[150, 200, 300]$ & - &  $[50, 100, 200]$\\
	& \textbf{Invariant CP-LSTM} & $[10, 25, 40]$ & $[150, 200, 300]$ &  $[30, 50, 100]$ & $[50, 100, 200]$\\
	& \textbf{TreeNet}* & $[0.001, 0.005, 0.008]$ & $[150, 200, 300]$ & - &  $[50, 100, 200]$\\
    \hline
    \hline
    \multirow{5}{*}{\rotatebox[origin=c]{90}{\textbf{TREC}}} & \textbf{Binary Sum-LSTM} & $[10, 25, 40]$ & $[150, 200, 300]$ & - &  $[0, 50, 100]$\\
	& \textbf{Binary CP-LSTM} & $[10, 25, 40]$ & $[150, 200, 300]$ &  $[30, 50, 100]$ &  $[0, 50, 100]$\\
    & \textbf{Child-Sum LSTM} & $[10, 25, 40]$ & $[150, 200, 300]$ & - &  $[0, 50, 100]$\\
	& \textbf{Invariant CP-LSTM} & $[10, 25, 40]$ & $[150, 200, 300]$ &  $[30, 50, 100]$ & $[0, 50, 100]$\\
	& \textbf{TreeNet}* & $[0.001, 0.005, 0.008]$ & $[150, 200, 300]$ & - &  $[0, 50, 100]$\\
\end{tabular}
\caption{Grids used for the model selection. The models with * validate the learning rate $lr$ rather than the batch size $bs$.} \label{tab:grid_search}
\end{table}

\end{document}